\documentclass{article}
\usepackage[main, final]{neurips_2026}

\title{\Large \textsf{QwenPaw-Data}: Bridging \emph{Facts}, \emph{Methodology}, and \emph{Execution} for Autonomous Enterprise Data Analytics \\ (Technical Report) }

\author{%
\mdseries
Tianjing Zeng$^\dag$, Yuntao Hong$^\dag$, %
Zhongjun Ding$^*$, Dandan Liu$^*$, Yinan Mei$^*$, \\%
Yunxiang Su$^*$, Yiming Wang$^*$, Xiaojian Zhang$^*$, Jingyu Zhu$^*$, Junhao Zhu$^*$, \\%
Zhuowen Liang, Jiazhen Peng, Lianggui Weng, Zhihao Ding, Kerui Yi, Qifeng Wang, \\%
Rong Zhu$^\ddag$, Bolin Ding$^\ddag$, Liyu Mou$^\ddag$, Jingren Zhou$^\ddag$ \\[4pt]%
{\bf Alibaba Group}%
}

\usepackage{color}
\usepackage{graphicx}
\usepackage{amsmath}
\usepackage{amsfonts}
\usepackage{listings}
\usepackage[ruled]{algorithm}
\usepackage[noend]{algorithmic}
\usepackage{url}
\usepackage[]{todonotes}%disable
\usepackage{verbatim}
\usepackage{xcolor}
\usepackage{multirow}
\usepackage{tabularray}
\usepackage{pifont}
\usepackage{xspace} 
\usepackage{tcolorbox}
\usepackage{colortbl}
\usepackage{subcaption}
\usepackage{makecell}
\usepackage{array}
\usepackage[table]{xcolor}
\usepackage{bbding, graphicx}
\usepackage{mdframed}
\usepackage{booktabs}
\usepackage{microtype}
\usepackage{fontawesome5}
\usepackage{float}
\usepackage{xurl}
\usepackage[hidelinks,breaklinks=true]{hyperref}

\newcommand{\dataaxon}{\textsf{\small QwenPaw-Data}\xspace}

\newcommand{\eg}{\textit{e.g.}\xspace}

\definecolor{darkgreen}{RGB}{0,170,0}
\definecolor{darkred}{RGB}{200,0,0}
\newcommand{\correct}{{\color{darkgreen}\ding{52}}\xspace}
\newcommand{\wrong}{{\color{darkred}\ding{55}\xspace}}
\newcommand{\pcorrect}{{\color{cyan}\ding{52}\rotatebox[origin=c]{-9.2}{\kern-0.7em\ding{55}\xspace}}}

\begin{document}
\maketitle

{\renewcommand{\thefootnote}{$^\dag$}\footnotetext{Co-first authors.}}
{\renewcommand{\thefootnote}{$^*$}\footnotetext{Equal contribution (listed in alphabetical order).}}
{\renewcommand{\thefootnote}{$^\ddag$}\footnotetext{Corresponding authors. Primary contact email: \texttt{red.zr@alibaba-inc.com} (Rong Zhu).}}

\begin{abstract}

Enterprise data analysis is emerging as a distinct frontier for autonomous agents. Compared with general-purpose interaction and software engineering, it operates in an open, ambiguous, and continuously evolving environment: business concepts must be grounded to the right entities, analytical procedures must be reproducible despite fuzzy feedback, and long-horizon workflows must execute over real enterprise data while preserving artifacts, provenance, and opportunities for human intervention. These characteristics call for a data-agent architecture that treats semantics, methodology, execution, and evolution as first-class system concerns.
To this end, we introduce \dataaxon, an agentic data system designed for enterprise intelligent data analysis. \dataaxon consolidates heterogeneous assets from warehouses, dashboards, documents, interaction logs, and historical tasks into reusable, governable, and evolvable analysis assets, then turns natural-language requests into end-to-end analytical workflows spanning data understanding, retrieval, analysis, report generation, and decision support.
Its architecture decomposes the problem into three collaborative subsystems: DataBridge provides trustworthy semantic grounding through interconnected metadata, knowledge, and trace graphs; Skill-Hub codifies expert analytical methodology into reusable and verifiable skills; and Host materializes these evidence and method assets into controllable, artifact-centric runtime execution. Across these subsystems, semantics, methods, traces, and feedback are continuously deposited back into the system, forming a self-evolving asset flywheel.
We deployed \dataaxon to serve business-intelligence users inside Alibaba, where it resolves objective queries accurately and earns far higher user satisfaction on open-ended analysis than a strong general-purpose agent. We further evaluate \dataaxon on public data-agent benchmarks for reproducibility, on which it consistently surpasses the current state of the art. Together, these results establish \dataaxon as a new paradigm for enterprise data analysis, pointing the way toward truly autonomous enterprise data agents.
\end{abstract}

\section{Introduction}
\label{sec:intro}

\subsection{Background}
\label{sec:intro-bkg}

In recent time, the rapid evolution of Large Language Models (LLMs) has profoundly driven the development of autonomous \emph{agents}, along with their underlying supporting frameworks, commonly referred to as \emph{harnesses}. Both general-purpose agents (\eg, chatting agents) and specialized agents (\eg, coding) have achieved remarkable progress in autonomy, expanded task scope, and engineering practicality. Modern coding agents have transcended the early limitations of generating isolated code snippets, acquiring the capability to independently manage task decomposition and debugging across the software development life cycle. Representative examples, such as OpenAI's CodeX~\citep{openai2026codex}, Anthropic's Claude Code~\citep{anthropic2026claudecode}, and frameworks like SWE-agent 2~\citep{DBLP:conf/nips/YangJWLYNP24}, demonstrate the ability to navigate complex codebases and resolve real-world issues. Currently, agent engineering has transitioned to scalable deployment, with industry surveys indicating that more than half of organizations have integrated agents into production environments~\citep{langchain2026state}.

In the realm of data science, \emph{data agents} have emerged as vital instruments for assisting data analysts (DA) and data scientists (DS). As an emerging ``blue ocean'' in the AI landscape, this field is increasingly attracting widespread attention for its capacity to drive profound business impact and strategic decision-making. Their primary technical directions focus on natural language interfaces (\eg, Text-to-SQL), automated data exploration, multi-step reasoning, and in-depth insight generation. For instance, frameworks such as Din-SQL~\citep{DBLP:conf/nips/PourrezaR23} and CHASE-SQL~\citep{DBLP:conf/iclr/PourrezaL0CTKGS25} leverage task decomposition and self-correction to optimize SQL generation, while end-to-end systems like Amazon QuickSight~\citep{aws2026quicksightgenbi} and Databricks Genie Agents~\citep{databricks2026genieagents} automate workflows from data ingestion to visual report generation. Researchers have substantially improved the accuracy of complex logical reasoning by incorporating lots of harness techniques, \eg, Chain-of-Thought prompting and multi-agent collaboration. These breakthroughs lower technical barriers and improve the efficiency of extracting actionable insights.

Despite the rapid proliferation of general-purpose, coding, and data agents, some fundamental distinctions and core boundaries among these paradigms remain insufficiently delineated.

\begin{enumerate}
    \item First, \textbf{\textit{what exactly are the fundamental distinctions and core boundaries that differentiate these paradigms from one another?}}

    \item Second, given these intrinsic characteristics and divergent operational environments, \textbf{\textit{is it genuinely necessary to establish a dedicated technological track exclusively for data agents, rather than merely adapting existing coding or general-purpose frameworks?}}
    
    \item Third, if such a distinct paradigm is indeed warranted, \textbf{\textit{what profound technical challenges must be overcome, and what specific enterprise demands should drive the tailored design of these specialized systems?}}
\end{enumerate}

In this technical report, we provide in-depth answers to these critical questions by systematically analyzing the unique characteristics and operational boundaries of enterprise data environments. Concurrently, we introduce our proposed solution, the \dataaxon system, which is specifically architected to overcome these profound technical challenges and fulfill the tailored demands of real-world data analytics.

\subsection{Data Agent: The Fundamental Difference}
\label{sec:intro-why}

A natural question is why powerful general-purpose and coding agents, backed by ever-larger foundation models, do not already solve enterprise data analysis. As the claim summarized in~\citep{anthropicdata}, \textbf{\emph{data analysis is not software engineering}}. We summarize the key differences in Table~\ref{tab:agent_comparison} and elaborate the details as follows.

\textbf{General-purpose agents are primarily designed as universal frameworks to tackle a broad spectrum of everyday tasks}, such as web navigation, operating system control, and general tool utilization. While they are capable of navigating dynamic, partially observable environments using multi-modal and heuristic feedback, they fundamentally lack the specialized mechanisms required for rigorous domain-specific problem-solving. Because they do not account for the stringent deterministic requirements of software engineering or the profound semantic ambiguity of enterprise data ecosystems, general-purpose agents are inherently ill-suited for complex coding and data analysis scenarios. Their generalized architectures fail to provide the domain-specific guardrails, structured verification loops, and deep contextual grounding necessary to ensure reliability and precision in these specialized fields.

\textbf{Coding agents primarily operate in a relatively deterministic, tool-supported verification environment}. In software engineering, mature programming toolchains---including compilers, unit tests, static analyzers, runtime execution, and logs---provide explicit and often binary feedback for diagnosing failures and refining implementations in a hard closed loop. Since coding tasks often admit multiple valid implementations that satisfy the same requirements and tests, this reliable feedback and high density of acceptable solutions make iterative search comparatively tractable.

\begin{table}[!t]
\centering
\caption{Comparative Analysis of General-Purpose, Coding, and Data Agents.}
\label{tab:agent_comparison}
\resizebox{\linewidth}{!}{
\renewcommand{\arraystretch}{1.12}
\begin{tabular}{@{}p{2.5cm}p{3.8cm}p{3.8cm}p{3.8cm}@{}}
\toprule
\textbf{Dimension} & \textbf{General-Purpose} & 
\textbf{Coding} & \textbf{Data} \\
\midrule
\textbf{Scenarios} & Web/OS control, tool use & Code generation, debugging & Business analytics \\
\textbf{Environment} & Dynamic, partially observable & Tool-supported, largely verifiable  & Open-space, ambiguous \\
\textbf{Feedback} & Multimodal, sparse, delayed & Explicit, binary, immediate & Fuzzy, subjective, human-reliant \\
\textbf{Solution Space} & Diverse trajectories & Diverse implementations & Fact-constrained conclusions \\
\textbf{Search Cost} & High (vast action space) & Light (tractable with verification) & High (strict constraints) \\
\textbf{Verification} & Weak closed-loop & Hard closed-loop & No deterministic proof \\
\addlinespace[0.25em]
\textbf{Required Guardrails} & Universal framework; lacks specialized guardrails & Syntax, logic, and execution verification & Semantic grounding, methodology, governance \\
\bottomrule
\end{tabular}
}
\end{table}

\textbf{Conversely, the data analysis tasks addressed by data agents are inherently \emph{open-space} problems}. Each step in the analytical workflow involves open-ended exploration without deterministic mechanisms to prove correctness. Unlike code compilation, data queries lack explicit, automated feedback signals, rendering the system heavily reliant on external validation, particularly human feedback. Compounding this challenge is the profound ambiguity in both task descriptions and intermediate results---such as concept-to-entity mapping and business context interpretation---which makes the requisite feedback signals inherently fuzzy and subjective. Moreover, in contrast to the diverse solutions in coding, data analysis often demands a single, unique correct answer derived from a single authoritative source to inform a specific business decision. This rigid uniqueness, coupled with a singular decision logic and the absence of deterministic guardrails, drastically increases the search cost and complexity for the model, making autonomous analytics significantly more arduous than code generation.

\textbf{\textit{In conclusion, traditional general-purpose and coding agents are fundamentally incapable of adapting to the intrinsic and stringent demands of data analysis scenarios. Therefore, the design and implementation of a dedicated data agent is not a mere elective, but an absolute imperative.}}

\subsection{Agents for Enterprise Data Analysis: The Difficulties}
\label{sec:intro-diff}

Next, we analyze the challenges confronting data agents that primarily stem from two fundamental dimensions. The first arises from the intrinsic characteristics of data analysis tasks themselves, which involve open-ended exploration, ambiguous feedback, and rigid solution uniqueness, as discussed above. The second originates from the complex nature of real-world enterprise data ecosystems, which are inherently decentralized, multi-modal, and continuously evolving.

\subsubsection{Difficulties from the Intrinsic Characteristics}
\label{sec:intro-diff-1}

To be more specific, the harness designed for a data agent requires deep business context, semantic layers, and governance. Consider a seemingly simple data analysis request: \emph{``analyze the average GAAP value of valid users for product X.''} We apply this simple question to explain the difficulties that a data agent needs to resolve:

\begin{itemize}
    \item \textbf{Grounding business concepts in the right data.}
    The request already hides two loaded terms, and most errors stem from failing to map such an ambiguous business concept onto the right data entity, which can break down in three compounding ways. 
    \begin{itemize}
        \item \emph{Concept--entity ambiguity}: ``valid users'' can resolve to different underlying entities---the registered-user table, the login-session table, or the paid-subscription table---each a legitimate mapping yet yielding a different number, while ``GAAP value'' must further be tied to the right revenue entity under the correct product scope.
        \item \emph{Staleness}: schemas and metric definitions change constantly, so unmaintained knowledge silently returns subtly wrong results. 
        \item  \emph{Retrieval failure}: even when the right entity exists and is well annotated, the vast search space keeps the agent from finding or using it correctly. The bottleneck is structured grounding, not access.
    \end{itemize}

    \item \textbf{Codifying analytical methodology.}
    Even with the data grounded correctly, the average GAAP value is still only a headline number. Answering the question well calls for a reproducible procedure---segmenting the metric across dimensions, checking distributions, and attributing anomalies---a form of procedural knowledge that can neither be retrieved as a single ``correct answer'' nor reliably internalized in model parameters. Such methodology must also be pitched at the right level of abstraction: overly generic methods fail to capture business intent, while overly specific ones are bound to a single domain and lose transferability. Moreover, since data analysis lacks compiler-like deterministic verification, a standardized, reusable, and verifiable analytical methodology becomes the structural safeguard that replaces the missing verification.
    \item \textbf{Sustaining long-horizon, artifact-centric execution.}
    Even with the right data and the right method, our example expands into a long chain of steps: querying the headline GAAP metric, detecting anomalies, decomposing and attributing it across dimensions, exploring new dimensions, and generating a report whose feedback is deposited for reuse. Such workflows are non-linear and long-running, directly execute SQL or Python over real business data, and produce reusable artifacts that must persist and be shared across sessions. This makes data analysis far more demanding than a short, self-contained action chain.
\end{itemize}

Unfortunately, as discussed in Section~\ref{sec:intro-why}, existing general-purpose and coding agents do not fully address the above difficulties:
\begin{itemize}
    \item General-purpose agents have no data-aware semantic model, so when pointed at a warehouse they must guess tables and fields (no grounding); they carry no codified analytical methodology (no methodology); and they are not designed for long-running, artifact-centered workflows (partial execution).
    
    \item Coding agents offer strong code generation and execution, so it is tempting to treat data analysis as a pure coding problem. Yet they too fall short on the same three dimensions: they can reach a schema but do not ground business semantics such as metric definitions (partial grounding); they encode no reusable analytical methodology (no methodology); and they run as a mostly linear action chain without long-horizon orchestration or cross-session artifact management (partial execution). As a result, a coding agent can produce runnable SQL yet still return a situationally wrong answer, and it cannot accumulate the analytical experience that future tasks depend on.
\end{itemize}

\subsubsection{Difficulties from the Enterprise Data Ecosystems}
\label{sec:intro-diff-2}

Despite the advancements, the design of existing data agents predominantly leans toward idealized and academic scenarios, rendering them insufficient for addressing practical enterprise challenges. In real-world corporate environments, data ecosystems are characterized by three fundamental complexities as follows:

\begin{enumerate}
    \item{\bf Decentralization.} Enterprise data is highly fragmented and multi-modal, scattered across implicit human expertise, knowledge documents, relational databases, and unstructured formats, rather than residing in centralized, clean schemas.

    \item{\bf Ambiguity and Uncertainty.} Analytical requests from stakeholders are inherently vague, necessitating a multi-step verification and interactive clarification process to progressively align with the \emph{exact} business intents.

    \item{\bf Growth and Evolution.} An effective analytical system must act as a growing entity that accumulates memory and self-evolves through fulfilled tasks, mirroring the cognitive development of human learning.
\end{enumerate}

Consequently, current static and narrowly scoped data agent systems fail to meet the dynamic, interactive, and continuous learning requirements of authentic enterprise data environments. Recent evaluations utilizing benchmarks designed to simulate realistic enterprise scenarios corroborate this deficiency. For instance, assessments on benchmarks like DABench~\citep{DBLP:journals/corr/abs-2603-20576} and AgenticDataBench~\citep{DBLP:journals/corr/abs-2607-01647} reveal that even state-of-the-art LLM-based agents still struggle significantly with complex, real-world data analysis tasks, falling far short of the performance required for practical deployment.

\subsection{\dataaxon: Our Solution}
\label{sec:intro-what}

To tackle the limitations of existing systems, we propose \dataaxon, an agentic data system built specifically for enterprise data analysis scenarios. 
\dataaxon consolidates the multi-source, multi-modal, and heterogeneous enterprise data, scattered across enterprise data warehouses, dashboards, business documents, interaction logs, and document knowledge bases, into reusable, governable, and evolvable analysis assets. Building on these assets, a user only needs to ask questions in natural language to interact with the agent to autonomously complete the end-to-end pipeline that spans data understanding, data retrieval, data analysis, report production, and ultimately decision support. At the same time, this system is strongly self-evolving, so every piece of user feedback and behavior helps it improve over time.

Unlike those systems, \dataaxon is natively designed and implemented to resolve these challenges. It reaches this not through a stronger model, but with delicate harness techniques to decompose each of the three difficulties onto a component. In addition, we carefully define the interaction logic between them, enabling them to collaborate smoothly to tune every analysis into reusable assets. This gives \dataaxon the following data-centric highlights that general-purpose and coding agents lack:

\begin{itemize}
\item{\textbf{Trustworthy Semantic Grounding System.}}
\dataaxon grounds every analysis in governed, graph-structured semantic evidence rather than prompt-level guessing, mapping ambiguous business concepts to the right data entities and eliminating data hallucination at its source.

\item{\textbf{Codified Analytical Methodology Hub.}}
\dataaxon supplies reusable, verifiable analytical skills that encode how an expert analyzes, so answers follow a reproducible methodology instead of ad-hoc, one-off reasoning.

\item{\textbf{Controllable Long-horizon Execution Environment.}}
\dataaxon orchestrates long-running, artifact-centric workflows with isolated SQL and Python execution, keeping every step traceable, verifiable, and open to mid-execution intervention rather than a restart.

\item{\textbf{Self-Evolving Data Assets.}}
Across all components in \dataaxon, semantics, methods, and experience are deposited as governable, evolvable assets, so the system keeps improving the more it is used.
\end{itemize}

With \dataaxon, enterprises can cover the serious analytical work a data team handles day to day, including but not limited to,

\begin{itemize}
\item \textbf{Business monitoring and anomaly diagnosis.} Track core metrics such as DAU, revenue, or conversion, and when one moves unexpectedly, automatically pinpoint which region, channel, or segment drove it.
\item \textbf{Trend and growth analysis.} Analyze the access, retention, or conversion trend of a product, identify turning points, and attribute the changes behind them.
\item \textbf{User and interaction insight.} Mine large volumes of dialogue and behavior logs to understand user intent, needs, and how they shift over time.
\item \textbf{Periodic business reporting.} Produce end-to-end monthly or quarterly reports for a product or business line, from data retrieval to a finished, decision-ready document.
\item \textbf{Ad-hoc deep-dive analysis.} Answer open questions such as ``why did product~X's revenue drop last month?'' with multi-dimensional decomposition and contribution attribution.
\end{itemize}

Across all of these, \dataaxon does not stop at returning a number: it grounds every metric in the right business definition, follows an expert analytical methodology, and delivers traceable, decision-ready results. By our evaluation (see Section~\ref{sec:eval}), \dataaxon substantially outperforms
strong general-purpose agents on open-ended tasks and public benchmarks.

\subsection{Comparison with Existing Systems}
To position \dataaxon within the current data-agent landscape, we also review and compare representative systems and commercial products across four key dimensions: semantic grounding, analytical methodology, execution capabilities, and self-evolution (discussed in Section~\ref{sec:intro-diff-1} and Section~\ref{sec:intro-what}).
Table~\ref{tab:system-comparison} presents a  comparison over representative industrial systems across these four dimensions.

\begin{table*}[t]
\centering
\caption{Comparison of representative data agent systems over four dimensions.}
\label{tab:system-comparison}

\small
\setlength{\tabcolsep}{4pt}
\renewcommand{\arraystretch}{1.25}

\resizebox{\linewidth}{!}{
\begin{tabular}{
    @{}
    >{\raggedright\arraybackslash}m{4.2cm}
    >{\centering\arraybackslash}m{2.4cm}
    >{\centering\arraybackslash}m{2.7cm}
    >{\centering\arraybackslash}m{2.5cm}
    >{\centering\arraybackslash}m{2.1cm}
    @{}
}
\toprule

\textbf{System}
&
\shortstack[c]{\textbf{Semantic}\\\textbf{Grounding}}
&
\shortstack[c]{\textbf{Analytical}\\\textbf{Methodology}}
&
\shortstack[c]{\textbf{Execution}\\\textbf{Capabilities}}
&
\shortstack[c]{\textbf{Self-}\\\textbf{Evolution}}
\\

\midrule

Oracle Analytics AI Agents
& \wrong
& \wrong
& \wrong
& \wrong
\\
\midrule

\shortstack[l]{Microsoft Fabric Data Agent}
& \correct
& \wrong
& \wrong
& \wrong
\\

\midrule

\shortstack[l]{QuickSight / Cognos / Intelligent Q / \\ BigQuery Analytics /  Volcengine Agent }
& \pcorrect
& \pcorrect
& \correct
& \wrong
\\
\midrule

\shortstack[l]{Databricks Genie Agent}
& \correct
& \correct
& \correct
& \pcorrect
\\

\midrule

\textbf{\dataaxon}
& \correct
& \correct
& \correct
& \correct
\\

\bottomrule
\end{tabular}
}
\vspace{3pt}

\begin{minipage}{\textwidth}
\footnotesize
\textit{Notation:}
\correct~explicit system-level support;
\pcorrect~partial, restricted, or insufficiently documented support;
\wrong~no explicit support or support limited to predefined functions.
\end{minipage}

\end{table*}

Oracle Analytics AI Agents~\citep{oracle2026analyticsaiagents} and Microsoft Fabric Data Agent~\citep{microsoft2026fabricdataagent} are comparatively assistant-oriented, but use different grounding mechanisms. Oracle configures agents through datasets, supplemental instructions, and knowledge documents, without exposing a standalone ontology or reusable skills. 
Microsoft Fabric Data Agent can use Fabric IQ Ontology~\citep{microsoft2026fabricontology} for governed semantic grounding; however, its standalone runtime remains query-centric, selecting data sources and generating read-only SQL, DAX, or KQL, with limited support for more complex planning and reusable analytical skills.

Amazon QuickSight~\citep{aws2026quicksightgenbi},
IBM Cognos Analytics Assistant~\citep{ibm2026cognosassistant},
Alibaba Quick BI Intelligent Q~\citep{alibaba2026quickbiintelligentq},
Google BigQuery Conversational Analytics~\citep{google2026bigqueryconversational},
and Volcengine Intelligent Analytics Agent~\citep{volcengine2026dataagent} represent a conversational-BI design center, supporting natural-language question answering, insight generation, dashboard construction, and report creation.
They typically rely on platform-native semantic assets---such as topics, semantic models, datasets, dashboard metadata, or warehouse governance objects---to translate business questions into BI queries and reports.
This design is effective for lowering the interaction barrier within a specific BI platform, but semantic grounding remains coupled to platform-local assets rather than being organized as a cross-system evidence substrate.
Analytical procedures are also exposed mainly through built-in BI capabilities, configured workflows, or prompt instructions, rather than reusable and verifiable method assets.

Databricks Genie Agents~\citep{databricks2026genieagents} is closest to \dataaxon in overall direction, combining an automatically maintained Genie Ontology~\citep{databricks2026genieontology} with multi-step analytical execution and reusable skills. 
However, while its ontology evolves continuously, its analytical skills do not support persistent self-evolution.

In contrast, \dataaxon provides explicit system-level support across all four dimensions, integrating governed semantic grounding, reusable analytical methodologies, controllable execution, and persistent self-evolution within a unified architecture.

\section{\dataaxon Architecture}
\label{sec:architecture}

\subsection{Design Principle}

The design principle of \dataaxon is to decompose enterprise data analysis according to the core questions an AI-native data agent must answer: what evidence to trust, what analytical method to follow, and how to execute the resulting workflow in a controllable way. A loose analogy can be drawn to how enterprise analysis teams collaborate across data modeling, BI (Business Intelligence) methodology, and execution infrastructure. 

This decomposition directly responds to the three difficulties identified in Section~\ref{sec:intro-diff-1}: grounding business concepts in the right data calls for an evidence-grounding subsystem; codifying analytical methodology calls for a method-orchestration subsystem; and sustaining long-horizon execution calls for an execution-control subsystem. Accordingly, \dataaxon decomposes an intelligent data analysis system into three decoupled modules with orthogonal responsibilities:

\begin{figure}[t]
\centering
\includegraphics[width=0.8\linewidth]{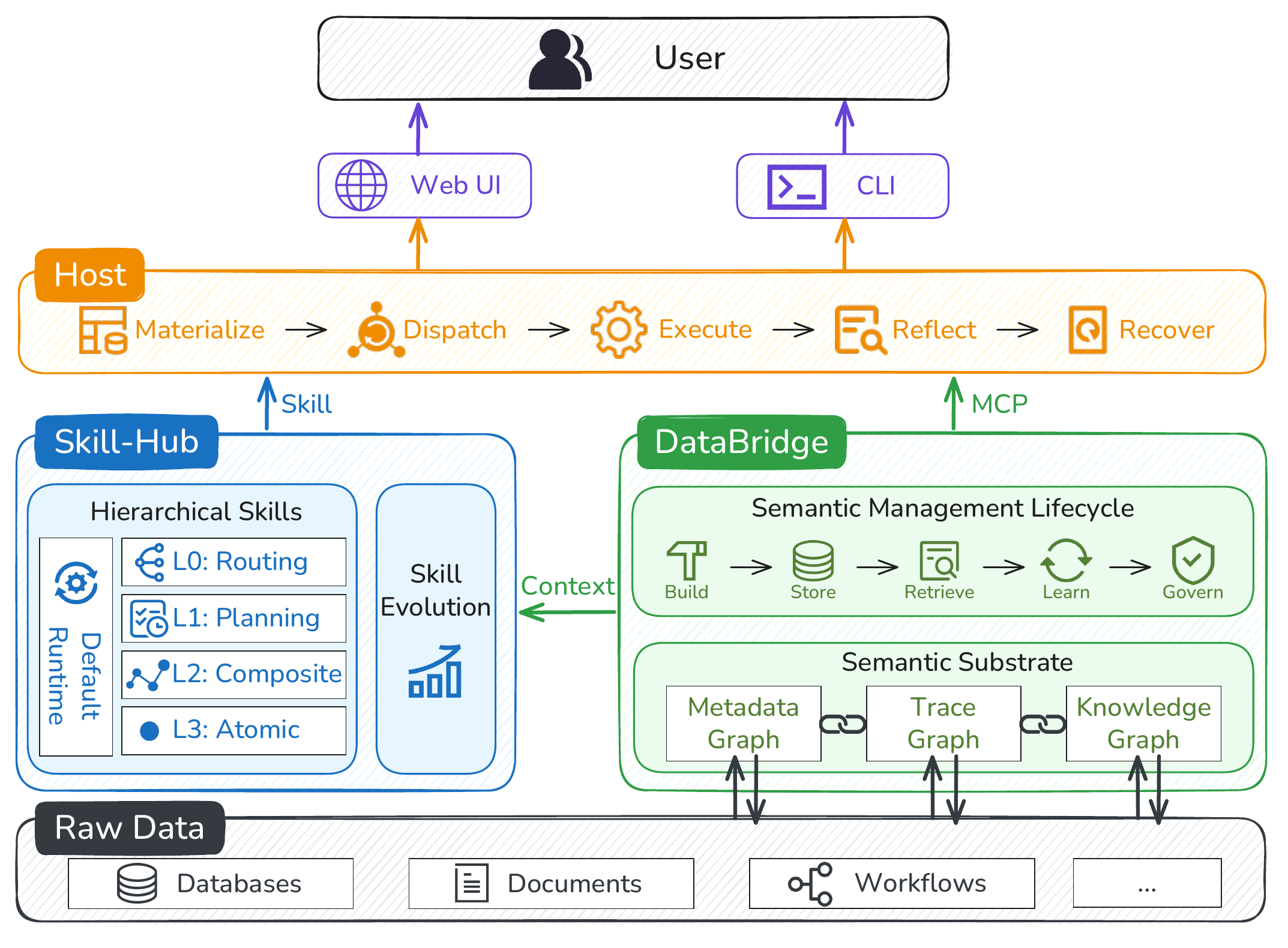}
\caption{The architecture overview of \dataaxon.}
\label{fig:architecture}
\end{figure}

\begin{itemize}
    \item \textbf{DataBridge} --- evidence grounding. It answers \emph{what facts to use}: it grounds business semantics, tracks staleness, and retrieves the minimum sufficient evidence for each task. Intuitively, it plays a role close to the semantic and modeling support provided by data warehouse teams.
    \item \textbf{Skill-Hub} --- method orchestration. It answers \emph{how to analyze}: it standardizes reusable analytical methods, workflows, procedural conventions, and quality expectations. Intuitively, it resembles the methodological and reporting expertise accumulated by BI teams.
    \item \textbf{Host} --- execution control. It answers \emph{how to run}: it materializes workflows as DAGs, isolates SQL and Python execution in sandboxes, invokes tools and subagents, and manages artifacts, traces, and recovery. Intuitively, it reflects the execution and infrastructure support usually provided by data/platform engineering teams.
\end{itemize}

Although many systems bundle evidence and methods into a single knowledge store, we keep DataBridge and Skill-Hub separate because they hold orthogonal knowledge: DataBridge captures volatile, domain-specific business facts (schemas, metrics, entities), while Skill-Hub captures stable, domain-general analytical methods. Merging them would bind each method to one domain, unable to transfer elsewhere or survive routine schema changes. Separating \emph{what facts to use} from \emph{how to analyze} instead lets one skill run over different evidence and lets evidence update without rewriting methods. Our ablation (Section~\ref{sec:eval-ablation}) confirms that removing either module causes a marked performance drop.

\subsection{System Overview}

Figure~\ref{fig:architecture} illustrates the overall architecture of \dataaxon. At the bottom, enterprise data assets are heterogeneous by nature: analytical tables, documents, searchable knowledge, historical traces, and operational data are scattered across different storage systems. DataBridge turns these scattered sources into a governed semantic substrate. Concretely, it organizes three graphs: the \emph{Metadata Graph (MG)} describes databases, tables, columns, metrics, dimensions, and lineage; the \emph{Knowledge Graph (KG)} captures business entities, definitions, rules, and organizational context; and the \emph{Trace Graph (TG)} records task traces, tool usage, intermediate artifacts, user feedback, and reusable experience. Downstream components do not retrieve raw fragments directly; they request task-specific semantic evidence from DataBridge.

Skill-Hub sits above this semantic substrate as the method subsystem. It organizes reusable skills into several categories. Routing skills identify user intent and dispatch it to the appropriate skills, and planning skills decompose a request into an executable analytical plan (a DAG). Workflow skills chain multiple atomic skills into end-to-end pipelines such as data fetching and metric analysis, while atomic skills package single reusable operations such as anomaly detection, dimensional drill-down, and report generation. Cross-cutting categories complement these, including runtime skills that encode execution-time behavioral guidelines and a meta skill that document how to use Skill-Hub itself. Importantly, Skill-Hub does not execute these skills by itself; it provides skill specifications, references, and scripts that Host can interpret and run. The same skills consume semantic evidence from DataBridge rather than encoding changing business facts inside the skill itself, which keeps analytical methods reusable across domains.

Host is the runtime subsystem that makes the semantic and methodological subsystems executable. It reads the skill specifications, references, and scripts provided by Skill-Hub, turns them into a DAG execution graph, schedules independent branches in parallel, exposes tools for SQL, Python, file operations, and report construction, runs tool calls inside isolated sandboxes, and maintains an artifact registry that links each intermediate result back to its producing action.

The three modules interact through two standard interfaces rather than ad hoc coupling: Host reaches DataBridge over the Model Context Protocol~\citep{mcp2024}, treating it as an MCP server that serves task-specific evidence on demand; and it consumes Skill-Hub through the open Agent Skills standard~\citep{agentskills2026}, which both Skill-Hub and Host adhere to, so the two sides align without custom integration code.

\subsection{An End-to-End Walkthrough}

\begin{figure}[!t]
\centering
\includegraphics[width=\linewidth]{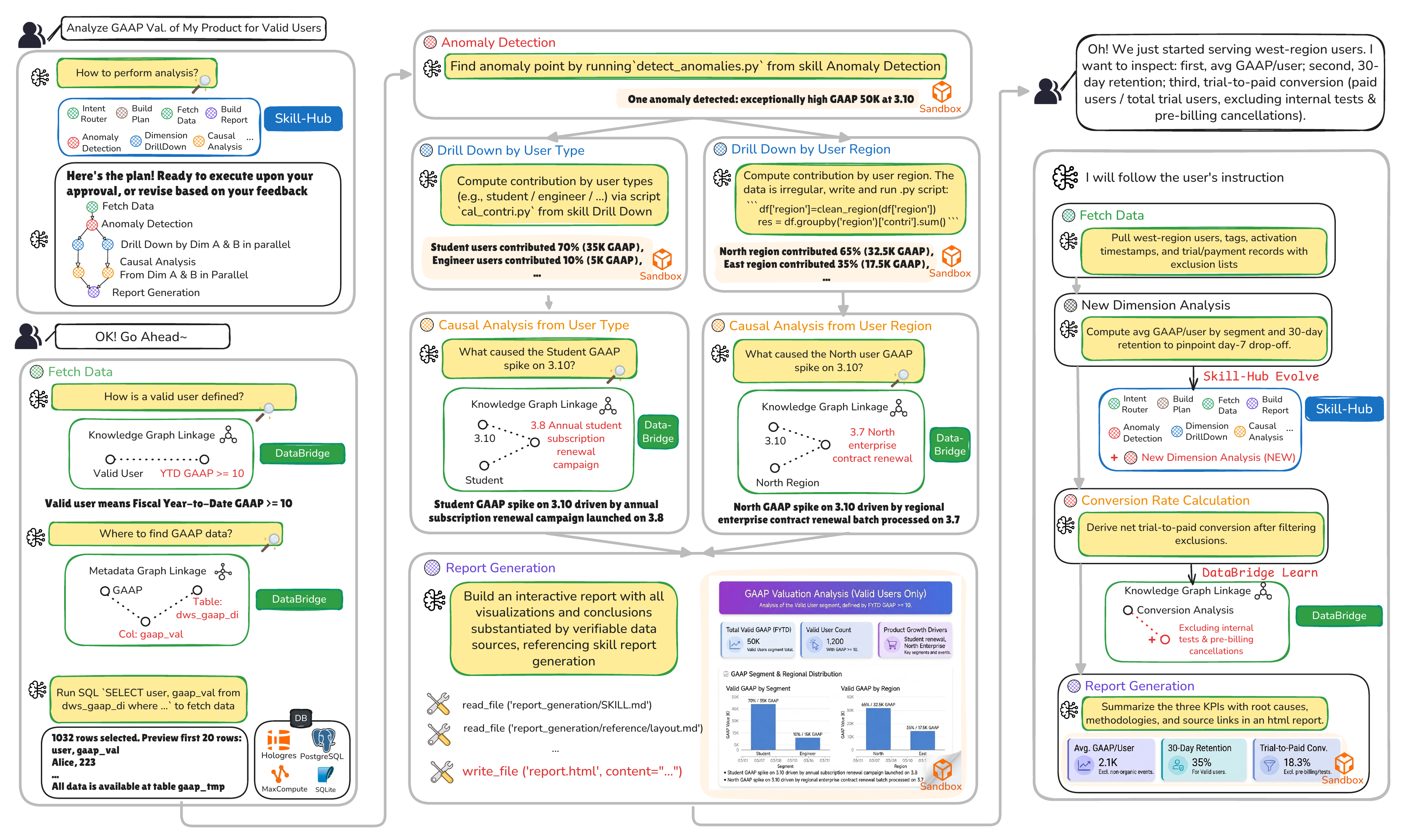}
\caption{Illustration of how \dataaxon empowers enterprise operations analytics.}
\label{fig:usecase}
\end{figure}

Figure~\ref{fig:usecase} traces a single request---\emph{``analyze the average GAAP value of valid users for product X''}---through the system, showing how DataBridge, Skill-Hub, and Host collaborate during a real analytical task.

\paragraph{Planning.}
The user first asks the question in natural language. Host consults Skill-Hub to select the relevant routing and planning skills, then uses their specifications and references to decompose the request into a DAG-shaped workflow: fetch data, detect anomalies, drill down by user type and region in parallel, conduct causal analysis, and generate the final report. Host materializes this plan as an executable DAG, keeps the plan inspectable, and waits for the user's approval or revision before running it. DataBridge can already participate at this stage by supplying high-level semantic hints that constrain what metrics, entities, and dimensions the plan should consider.

\paragraph{Data retrieval.}
Once execution starts, the first question is not how to write SQL, but what the business terms mean. DataBridge resolves ``valid users'' through the Knowledge Graph and locates the GAAP metric through the Metadata Graph, linking the logical definition to the physical table and column. Host then invokes the corresponding data-access tools and executes the generated SQL in a sandboxed environment against the underlying warehouse. The resulting dataset is registered as an artifact, so later conclusions can be traced back to the exact query, source, and previewed rows.

\paragraph{Analysis.}
During analysis, Host follows the reusable skill assets from Skill-Hub, including scripts and references for anomaly detection, dimensional drill-down, contribution calculation, and causal analysis. Host schedules independent branches of the DAG in parallel, for example, user-type and region drill-downs can run concurrently, and runs their scripts and tool calls inside an isolated sandbox environment. During attribution, DataBridge continues to provide grounded evidence from KG, linking observed metric changes to business events rather than letting the model guess from prompt context alone.

\paragraph{Report generation.}
After the analytical branches converge, Host uses the report-generation skill package from Skill-Hub to organize findings, charts, and methods into an interactive report. Each chart and conclusion in the report links back to its underlying data source, which can be downloaded and audited.

\paragraph{Self-evolution.}
Beyond delivering a report, each completed task feeds back to evolve the system. In our example, the user follows up with a new analysis demand, such as adding a new dimension, and confirms refined metric definitions. Host executes these follow-up instructions and captures the resulting execution trace. These signals then feed back into the two asset subsystems: Skill-Hub turns the newly requested procedure into a candidate reusable skill, and DataBridge promotes the confirmed definitions and validated semantic corrections into KG. In this way, a completed task becomes reusable experience for the next similar analysis.

\begin{figure}[t]
\centering
\includegraphics[width=\linewidth]{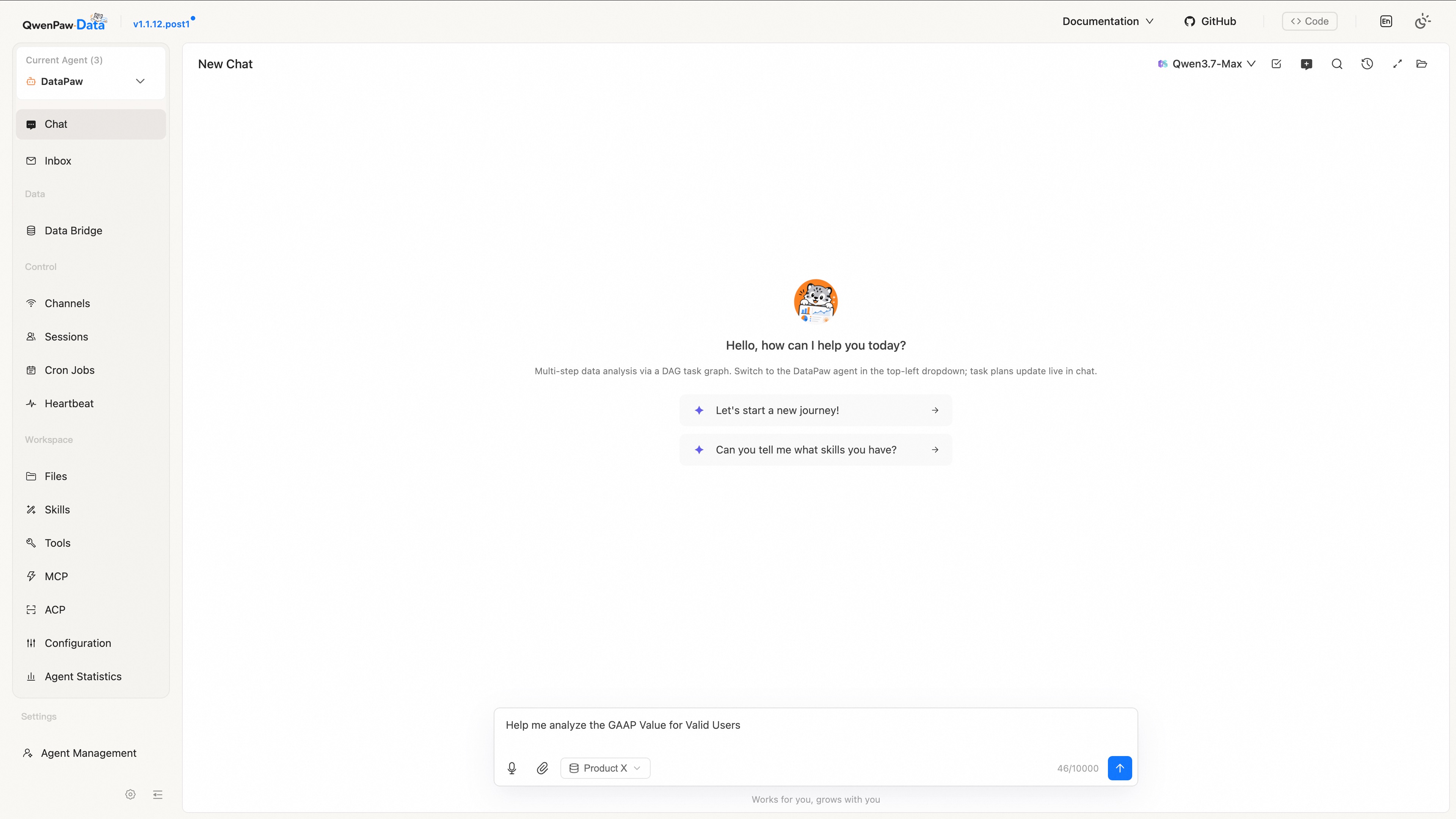}
\caption{The ChatWeb interface of \dataaxon, delivered as a plugin on top of \textsf{QwenPaw}.}
\label{fig:webui}
\end{figure}

\subsection{How to Use \dataaxon}

\dataaxon provides two usage modes: ChatWeb and CLI/SDK. ChatWeb targets interactive business analysis, while CLI/SDK targets platform integration and secondary development.

\paragraph{ChatWeb.}
In ChatWeb mode, \dataaxon is delivered as a plugin on top of \textsf{QwenPaw}~\citep{qwenpaw}, a general-purpose agent platform, and reuses its conversational frontend, so an existing chat interface is extended with data-analysis capabilities without rebuilding the interaction layer. The frontend is further adapted for data workloads, for example to visualize the parallel DAG task plan and to track long-running, multi-step analyses, as shown in Figure~\ref{fig:webui}. Through ChatWeb, users submit analytical requests in natural language, inspect the generated plan, track task progress, view intermediate artifacts, and continue refining the analysis in the same conversation. This mode suits business users and analysts who need an interactive data-analysis experience.

\paragraph{CLI/SDK.}
CLI/SDK mode is designed for platform integration and secondary development. \dataaxon exposes its core capabilities, including intent understanding, task planning, and workflow execution, through a lightweight command-line interface, so it can be embedded into existing enterprise BI platforms without a dedicated web frontend. Developers can extend data sources, register new tools, customize analytical skills, and define domain-specific semantic rules. As a result, \dataaxon can be seamlessly embedded into existing enterprise data platforms, internal BI systems, workflow engines, or other service environments, serving as an infrastructure layer for building customized data agents.

\section{DataBridge}
\label{sec:databridge}

\subsection{Role: The Semantic Evidence Subsystem}

DataBridge is the semantic evidence subsystem of \dataaxon. Its responsibility is to turn fragmented enterprise data, business knowledge, and historical task experience into governed evidence that Host can retrieve and use during analytical execution. This is the component that supports trustworthy semantic grounding: instead of asking the model to infer from prompt-level context, DataBridge exposes verified, traceable, and governable evidence for each task.

This role is necessary because enterprise analysis is an open-form task. Business concepts are ambiguous, metric definitions change over time, and useful experience is scattered across past tasks, documents, data models, and user feedback. DataBridge therefore does not behave like a generic text retriever. It maintains a semantic evidence substrate that can be built from heterogeneous sources, stored with provenance and validity, retrieved as task-specific context, improved from feedback, and governed when definitions become stale or conflicting.

\begin{figure}[t]
\centering
\includegraphics[width=0.8\linewidth]{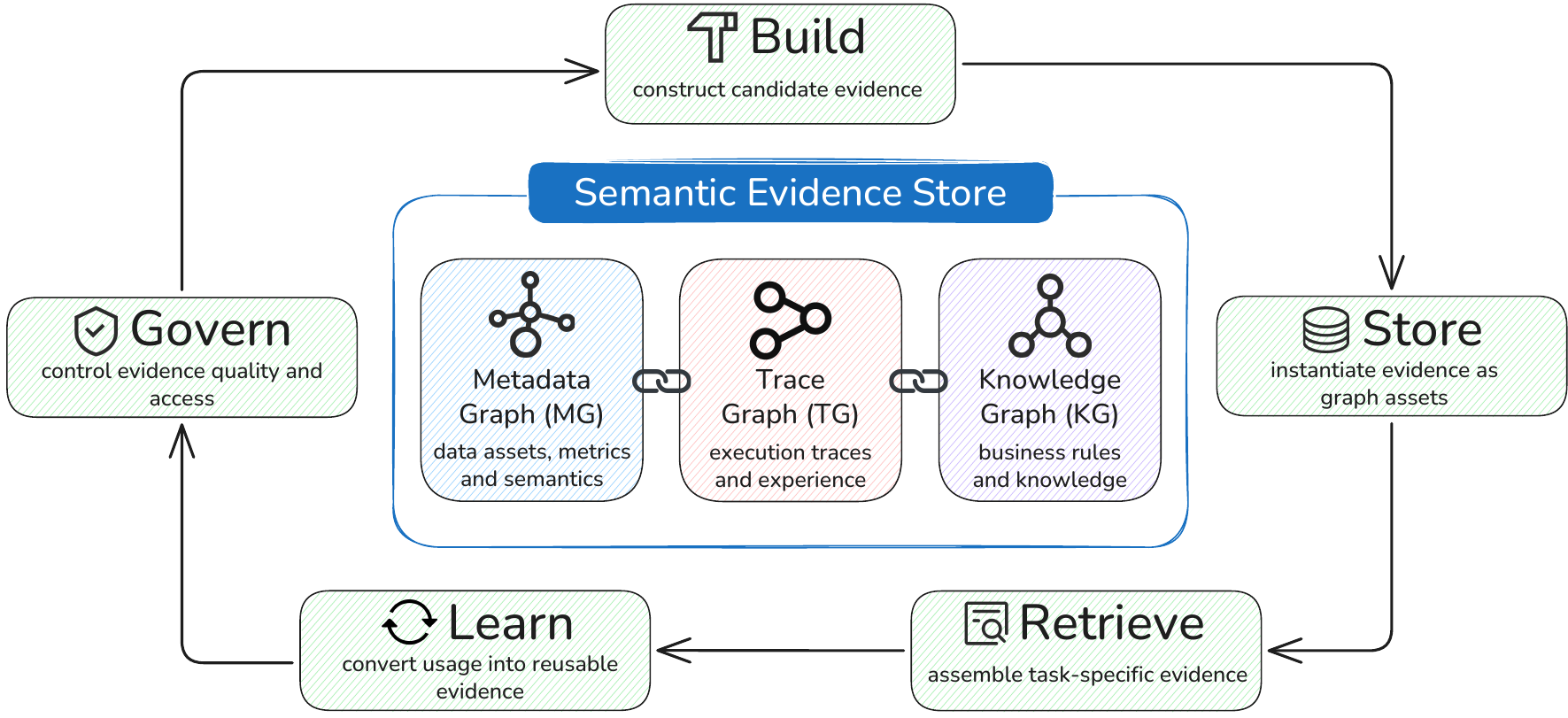}
\caption{The overall structure of DataBridge. The center shows the semantic evidence stores, while the outer loop shows how evidence is built, stored, retrieved, learned, and governed.}
\label{fig:databridge}
\end{figure}

Figure~\ref{fig:databridge} provides the high-level map of DataBridge. The center represents the semantic substrate that stores different kinds of evidence, while the surrounding lifecycle explains how evidence enters the system, becomes retrievable, returns to Host during task execution, and evolves through feedback and governance. The rest of this section follows this logic: we first introduce what kinds of information DataBridge manages, then explain how the lifecycle operates over them.

\subsection{What DataBridge Manages}

Before introducing the storage structure, it is useful to clarify what kinds of information DataBridge manages. In enterprise analysis, the evidence needed by an agent is not a single document or table schema, but a combination of three information types.

\paragraph{Data semantics.}
Data semantics describes the structure and meaning of data assets: databases, tables, columns, metrics, dimensions, lineage, partitions, and data-source relationships. In the GAAP use case of Figure~\ref{fig:usecase}, this information tells Host where the GAAP metric is physically stored, how it maps to the underlying table and column, which product and time fields must be used, and which analytical dimensions such as user type or region are available for drill-down.

\paragraph{Business knowledge.}
Business knowledge captures concepts and rules that are not fully encoded in schemas. It includes business entities, metric definitions, operational rules, policies, events, and external definitions. In the same use case, this is where DataBridge represents the definition of ``valid users,'' GAAP-related business rules, and business events such as subscription-renewal campaigns that may explain metric changes.

\paragraph{Task experience.}
Task experience records what has happened in previous analytical tasks: execution traces, tool calls, intermediate artifacts, successful analysis paths, failed attempts, user corrections, and reusable lessons. For recurring GAAP analyses, this information helps the system remember which drill-down branches were useful, which scripts or tool calls produced reliable evidence, which assumptions users corrected, and which historical traces should be reused during attribution.

\subsection{Semantic Evidence Lifecycle}

\begin{figure}[t]
\centering
\includegraphics[width=\linewidth]{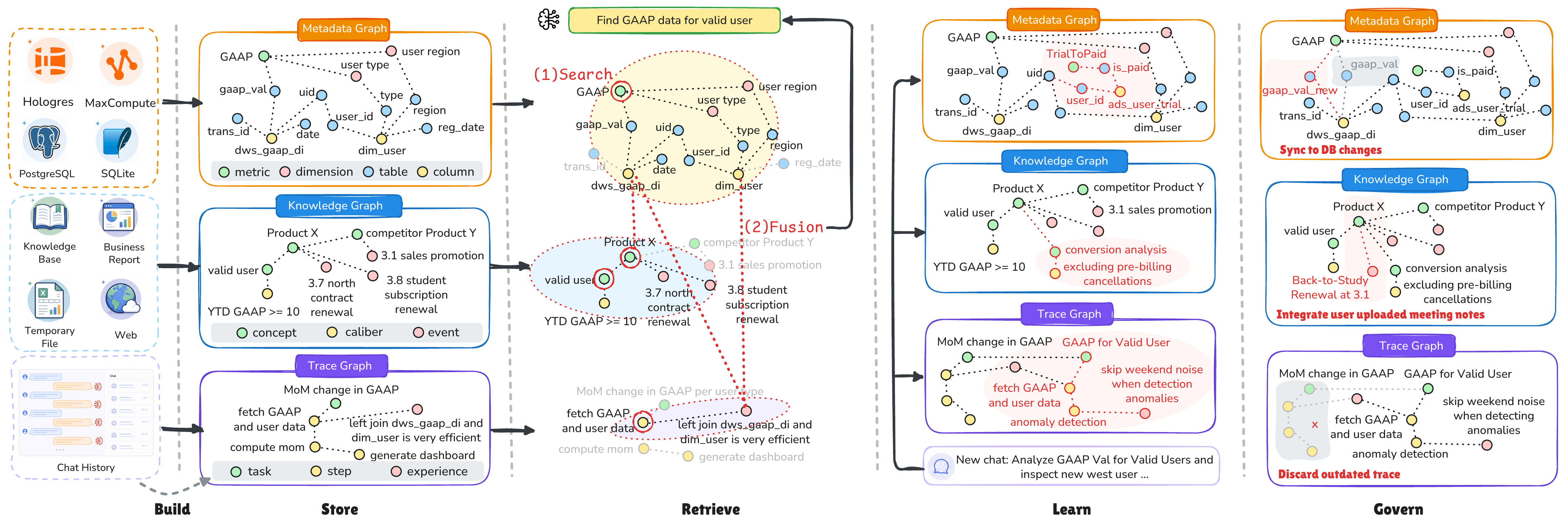}
\caption{The semantic evidence lifecycle of DataBridge on the GAAP use case, spanning five stages: Build ingests and normalizes heterogeneous sources, Store instantiates them as three semantically interconnected graphs, Retrieve assembles task-specific evidence via cross-graph search and fusion, Learn distills new metrics, rules, and experiences from usage, and Govern keeps the evidence reliable.}
\label{fig:databridge-case}
\end{figure}

DataBridge manages the three information types through a five-stage lifecycle: Build, Store, Retrieve, Learn, and Govern. The lifecycle is designed to separate evidence construction from evidence use: information is first collected and normalized, then instantiated as governed graph assets, retrieved as task-specific context, improved from usage, and continuously controlled before it can influence future analytical execution. Figure~\ref{fig:databridge-case} illustrates this lifecycle with the GAAP analysis example, which we follow throughout this subsection.

\subsubsection{Build: Construct Candidate Evidence}

The Build stage converts heterogeneous enterprise sources into candidate semantic evidence. Its input includes structured metadata from warehouses and metric systems, semi-structured knowledge from documents and dashboards, and behavioral signals from historical tasks, reports, intermediate artifacts, user feedback, and failure records. These sources are collected together because enterprise analysis usually depends on evidence scattered across systems rather than on a single schema or document.

The main function of Build is normalization. DataBridge extracts candidate entities, relations, definitions, and traces from raw sources, aligns them to the three information types, and attaches source information and preliminary confidence signals. At this stage, the output is not yet trusted graph knowledge; it is candidate evidence prepared for graph storage and later governance. In the GAAP use case, Build may extract a candidate GAAP metric from warehouse schemas, capture the business concept of ``valid user'' from business reports and knowledge bases, and mine reusable experience from chat history, such as the observation that left joining \texttt{dws\_gaap\_di} with \texttt{dim\_user} is efficient. Together, these candidates later help the LLM understand what ``GAAP for valid users'' actually refers to, where the underlying data lives, and how to retrieve it efficiently, so that similar analyses can reuse this evidence instead of rediscovering everything from scratch.

\subsubsection{Store: Instantiate Evidence as Graph Assets}

The Store stage turns candidate evidence into persistent graph assets. It does not simply archive the three information types; it instantiates them as typed nodes, typed edges, and cross-graph links that can be retrieved and governed:
\begin{itemize}
    \item \emph{Metadata Graph (MG)} represents data assets as schema-level nodes, such as databases, tables, columns, metrics, dimensions, and partitions. Edges encode structural and semantic relations, such as table--column containment, metric--column mapping, metric--dimension compatibility, and lineage between upstream and downstream assets.
    \item \emph{Knowledge Graph (KG)} represents business knowledge as concept-level nodes, such as business entities, metric definitions, rules, policies, events, and external concepts. Edges encode relations such as definition, constraint, scope, temporal applicability, and links from business concepts back to the MG assets that implement them.
    \item \emph{Trace Graph (TG)} represents task experience as execution-level nodes, such as tasks, plans, tool calls, intermediate artifacts, reports, feedback, and corrections. Edges encode how an analysis was executed, including which action produced which artifact, which evidence was used, which branch succeeded or failed, and which user correction updated MG or KG evidence.
\end{itemize}
Each node and edge is stored with governance metadata, including provenance, trust level, temporal validity, lifecycle state, and visibility. This metadata is part of the graph representation itself, so downstream retrieval can distinguish verified, outdated, conflicting, or restricted evidence. In the GAAP use case, Store turns these candidates into typed nodes and edges, each supporting a later step. In MG, the \texttt{dws\_gaap\_di} table contains the \texttt{gaap\_val} column that the GAAP metric maps to, which grounds data retrieval, and the metric stays compatible with the user type and region dimensions used for drill-down analysis. In KG, the ``valid user'' calibration \emph{YTD GAAP $\geq$ 10} fixes the filter applied when fetching data, while renewal events such as the student-subscription campaigns feed the later causal analysis. In TG, the prior task, its steps, and its artifacts, together with the reusable experience distilled by the Learn stage (Section~\ref{sec:databridge-learn}), speed up a follow-up similar request with the efficient join it already found.

\subsubsection{Retrieve: Assemble Task-Specific Evidence}

The Retrieve stage assembles the minimum sufficient evidence needed by a concrete task. Given the user request, current plan, and session state from Host, DataBridge anchors task mentions to graph nodes instead of retrieving raw text chunks. It then expands task-specific subgraphs across MG, KG, and TG, fuses the selected evidence, and returns a structured context package or concise natural-language answer back to Host.

The design goal of Retrieve is precision under governance. MG contributes concrete data assets and valid analytical dimensions; KG contributes business definitions, rules, scopes, and events; TG contributes prior execution experience, reusable paths, and known pitfalls. Retrieve combines these signals while respecting provenance, validity, visibility, and conflict status, so Host receives evidence that is compact enough to act on and reliable enough to ground execution. In the GAAP use case, for a request such as ``find GAAP data for valid users,'' Retrieve anchors three mentions to graph nodes: the GAAP metric in MG, the ``valid user'' concept together with the current product in KG, and the prior fetch-data step in TG. Each anchor then expands outward into the semantically related context around it, and the resulting subgraphs are fused across layers, for instance by linking the \texttt{dws\_gaap\_di} and \texttt{dim\_user} tables in MG to the join experience carried by that fetch-data step in TG. Together they assemble a compact yet complete context, so Host works from grounded, traceable evidence.

\subsubsection{Learn: Convert Usage into Reusable Evidence}
\label{sec:databridge-learn}

The Learn stage turns real analytical usage into reusable evidence. It has two complementary paths: online learning absorbs explicit feedback during task execution, such as user-confirmed definitions, corrected rules, rejected assumptions, and report review comments; offline learning mines accumulated traces for recurring patterns, stable metric--dimension relations, useful drill-down paths, and verified failure modes.

The output of Learn is a set of candidate updates to MG, KG, or TG. These updates are intentionally separated from direct execution. A correction may suggest a KG rule update, a repeated successful workflow may suggest a TG experience pattern, and a verified metric--column relation may strengthen an MG mapping. This preserves the boundary between DataBridge and Skill-Hub: DataBridge evolves semantic evidence, traces, and rules, while Skill-Hub evolves reusable method assets such as skill packages, references, and scripts. In the GAAP use case (Figure~\ref{fig:usecase}), Learn works on the execution trace and explicit user feedback accumulated across the whole analysis, distilling three candidate updates. The user's definition of net conversion becomes a new trial-to-paid conversion metric in MG, along with its supporting columns and table. A correction the user made turns into a KG rule \emph{excluding pre-billing cancellations}. The successful run, meanwhile, leaves TG a reusable experience, \emph{skipping weekend noise during anomaly detection}. Once promoted, DataBridge supplies these updates as evidence for later requests, so the agent automatically conducts subsequent analyses under the same definition, exclusion, and caveat; this reduces the attention the user must devote to restating them and improves the efficiency of recurring analyses. Until promotion, they remain candidates rather than one-off session notes.

\subsubsection{Govern: Control Evidence Quality and Access}

The Govern stage keeps the evidence layer reliable over time. It controls whether graph evidence is allowed to affect retrieval and execution. DataBridge checks provenance, temporal validity, conflicts, trust level, lifecycle state, and visibility; promotes validated candidate updates; deprecates stale nodes; and prevents conflicting or unauthorized evidence from being used silently.

Govern is both an automated control process and a human-facing correction process. DataBridge exposes visual governance interfaces for MG, KG, and TG, allowing users to inspect the three graphs and directly create, update, or delete graph nodes, edges, rules, and trace records when human correction is required. These operations update the governed evidence layer rather than patching a single prompt or session. In the GAAP use case, Govern updates, adds, and removes evidence to keep the three graphs aligned with reality. When an upstream schema change in the \texttt{dws\_gaap\_di} table renames the GAAP column from \texttt{gaap\_val} to \texttt{gaap\_val\_new}, it updates MG by deprecating the old field and redirecting to the new one, so that retrieval never binds to a dropped column. Also, in response to user-uploaded meeting notes that surface a previously unrecorded campaign, it adds a new \emph{Back-to-Study Renewal} event to KG on March 1. And as time passes, a trace or experience in TG that is no longer used or whose definition no longer holds is removed before it can mislead future attribution. Through these operations, only current and trustworthy evidence reaches Retrieve, and superseded or conflicting evidence never silently influences execution.

\subsection{What DataBridge Guarantees}

Through the interaction of the three information types and the five-stage lifecycle, DataBridge provides three guarantees for \dataaxon:
\begin{itemize}
    \item \emph{Groundedness}: business concepts are mapped to verified graph evidence rather than inferred from prompt context.
    \item \emph{Freshness}: definitions, schemas, and business rules carry validity and lifecycle state, so stale evidence can be detected and governed.
    \item \emph{Evolvability}: user feedback, execution traces, and confirmed corrections become reusable semantic assets for future tasks.
\end{itemize}
Together, these guarantees allow Host to execute analytical workflows over governed evidence rather than over fragmented, unverified context.

\section{Skill-Hub}
\label{sec:skillhub}
\subsection{Role: The Method Asset Subsystem}

Skill-Hub is the method asset subsystem of \dataaxon. Its responsibility is to turn reusable analytical methodology into structured skill assets that Host can interpret and execute during data-analysis tasks. Skill-Hub answers \emph{how to analyze}: the analytical method a task should follow, the operations each step involves, the intermediate artifacts it should produce, and the criteria for checking the result.

This role is necessary because sound data analysis relies on expert methodology that a model cannot reliably reinvent for every task. Skill-Hub therefore packages this methodology as structured assets, including skill specifications, references, and scripts. It does not execute these assets itself but leaves their interpretation and execution to Host. Because each asset carries explicit inputs, outputs, expected artifacts, and evaluation criteria, it also serves as a soft verification loop that partially compensates for the lack of compiler- or test-based feedback in data analysis. 

\begin{figure}[t]
\centering
\includegraphics[width=0.7\linewidth]{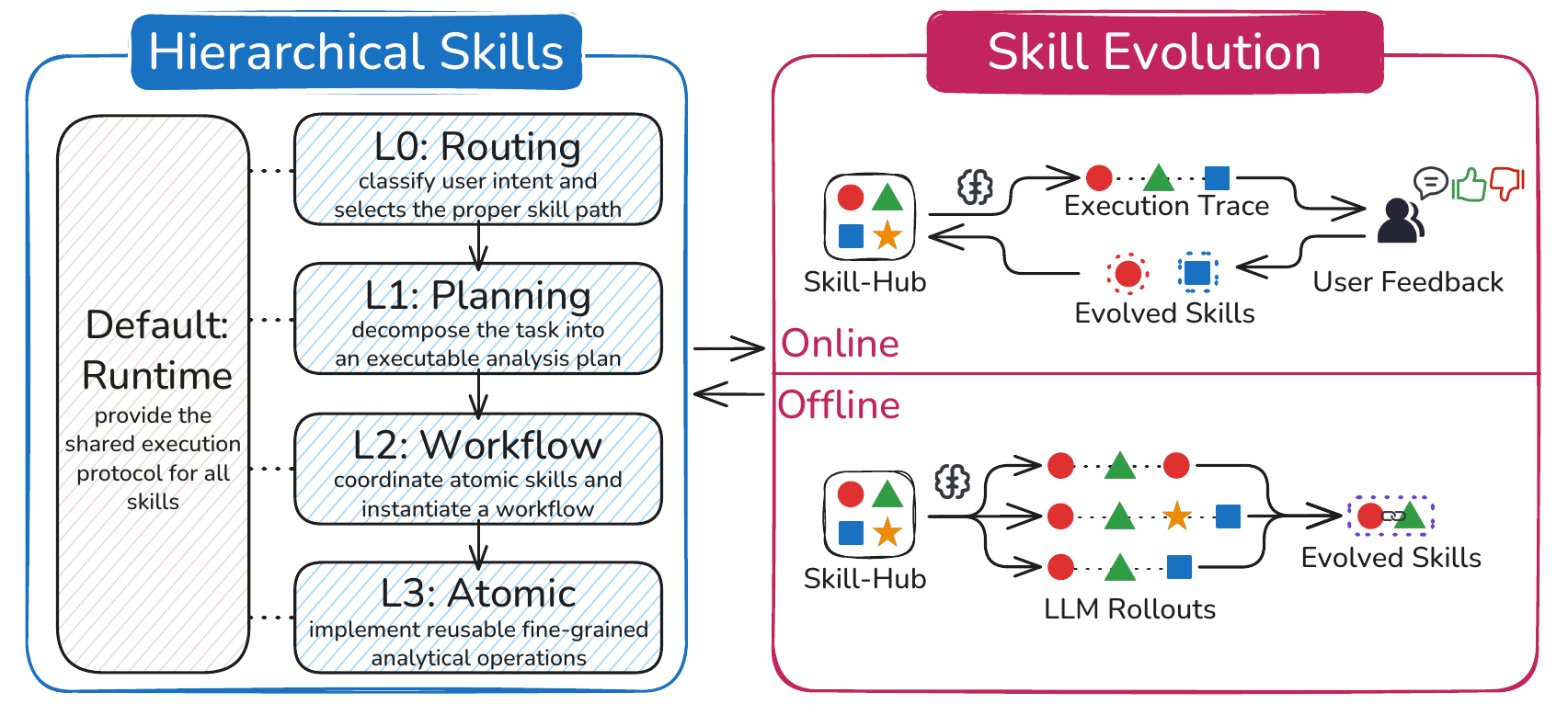}
\caption{Hierarchical skills and skill evolution in Skill-Hub. The left part shows the skill hierarchy and the right part shows the offline and online evolution loops that turn traces and feedback into versioned skill updates.}
\label{fig:skillhub}
\end{figure}

To fulfill this role, Skill-Hub is built along two axes, summarized in Figure~\ref{fig:skillhub}: a hierarchy that decomposes analytical methodology from task routing down to atomic operations, and an evolution loop that keeps these method assets updated through evaluation, historical traces, and user feedback. The rest of this section describes the assets Skill-Hub manages, this hierarchy, and how skills are validated and evolved.

\subsection{What Skill-Hub Manages}

The basic unit Skill-Hub manages is a skill, which is more than a single prompt or script. Following the Agent Skills standard~\citep{agentskills2026}, each skill is packaged from a \emph{skill specification}, \emph{references}, and \emph{scripts}, together forming a reusable method package that Host can interpret and that DataBridge can ground with evidence.

\paragraph{Skill specifications.}
Skill specifications, typically written in \texttt{SKILL.md}, define the structured contract and procedural logic of an analytical method. They describe the task types a skill applies to, the semantic evidence it expects from DataBridge, the inputs Host should provide, and the outputs and artifacts that should be produced. More importantly, this is where the analysis workflow is specified: the procedure steps, step dependencies, branching conditions, revision points, and quality expectations are all described at the method level. Host then interprets this specification, materializes the workflow as an executable DAG, and manages the actual execution.

\paragraph{References.}
References hold the detailed material that individual skill steps rely on, such as report layouts, visualization conventions, or the analysis methods used for different business types. Keeping this detail out of the skill specification enables progressive disclosure: Host loads a reference only when the corresponding step needs it, instead of injecting all method detail up front and diluting the LLM's attention. 

\paragraph{Scripts.}
Scripts package standard computations, such as common statistical routines, metric calculations, and report helpers, into ready-to-call functions rather than code the model regenerates each time. This keeps recurring operations standardized and stable across tasks and saves tokens, since Host can invoke a tested function instead of synthesizing and reasoning through the code again. As with the other assets, Skill-Hub only provides these scripts, and Host is responsible for executing them.

\subsection{Hierarchical Skill Organization}

Skill-Hub organizes method assets into a hierarchy so that complex analytical workflows can be composed from reusable parts. The hierarchy moves from coarse-grained task understanding down to fine-grained analytical operations, and is complemented by cross-cutting skills that apply throughout a task. 

\paragraph{L0: Routing.}
Routing skills classify a user request into a task family. Common task families include metadata query, data query, BI business analysis, data exploration, statistical modeling, and report generation. The output is a task-family decision that Host uses to select the relevant downstream skills when constructing the executable workflow.

\paragraph{L1: Planning.}
Planning skills decompose a task into an inspectable analytical plan. This typically includes identifying the analysis scenario, refining the required metrics, choosing analysis perspectives, and specifying which evidence to request from DataBridge, guided by references such as scenario-type and metric specifications. The output is a plan specification rather than an executed plan; Host materializes it as a DAG and manages user inspection, revision, and execution.

\paragraph{L2: Workflow.}
Workflow skills chain multiple atomic skills into an end-to-end pipeline for a task family, defining the sequence of steps and their dependencies. Representative workflows are BI analyses such as metric, cohort, conversion, and retention analysis, with additional pipelines covering supporting tasks like data fetching and modeling. Host usually materializes each pipeline into executable DAG actions.

\paragraph{L3: Atomic.}
Atomic skills are the smallest reusable analytical operations and serve as building blocks for workflows. They cover recurring operations such as adaptive thresholding and anomaly detection, dimensional drill-down, attribution and causal analysis, clustering, comparison, conversion- and retention-rate computation, distribution and funnel analysis, LTV analysis, and dashboard and report generation. Because each atomic skill carries only method, different workflows can reuse the same skill with different evidence supplied by DataBridge.

\paragraph{Cross-cutting skills.}
Some skills apply across the hierarchy rather than at a single level. Runtime skills encode execution-time behavioral guidelines, such as user-interaction strategy and semantic-layer usage, that shape how any skill behaves during a task. Meta skills document how to use the skill system itself and are loaded before a task begins, so the agent knows how to navigate and combine the other skills.

This hierarchy balances consistency with reuse: high-level skills keep the overall analytical path consistent, while atomic skills make common operations reusable across workflows. Because business facts remain in DataBridge, the same method hierarchy can be reused across different data domains.

\subsection{Skill Validation}

Skill-Hub provides methodological control for open-form data analysis. Because most data analysis has no deterministic test to confirm correctness, some form of methodological check is still needed. Rather than hard-coding a fixed checklist into each skill, which would be rigid and incomplete, \dataaxon automatically derives a task-specific checklist from the skill specification, references, expected artifacts, and current task context. These generated checklists make the analytical process inspectable without turning the skill itself into a static list of checks.

At execution time, a reflection subagent runs against this checklist, examining whether the required evidence has been retrieved, whether key assumptions have been stated, whether the expected intermediate artifacts have been produced, whether the selected dimensions match the task intent, and whether the final report links its conclusions back to supporting artifacts. Such validation does not prove that a business conclusion is absolutely correct; rather, it confirms that the intended method was followed in a reproducible and reviewable way. This gives Skill-Hub a soft verification loop that partially compensates for the absence of compiler-like feedback in data analysis.

\subsection{Skill Evolution}

Skill-Hub evolves continuously so that analytical methods keep improving with real usage, forming the method half of the data-asset flywheel alongside DataBridge: DataBridge accumulates semantic evidence, traces, and rules, while Skill-Hub accumulates reusable methodology, workflow structures, references, and scripts. Evolution draws on two sources of signal, offline traces and online feedback, but neither changes an active skill directly; both only produce candidate revisions that become active through governed, versioned promotion.

\paragraph{Offline evolution.}
Offline evolution mines benchmark tasks, historical cases, and rollout traces. Stable patterns across successful cases surface reusable workflows, better decomposition strategies, or common quality checks, while divergent or failed traces expose ambiguous decisions, missing constraints, or failure-prone method choices. Both are collected as candidate skill revisions.

\paragraph{Online evolution.}
Online evolution draws on human-in-the-loop feedback during real tasks, where users may correct a procedure, request an additional perspective, reject an assumption, or ask for a different report structure. Such feedback becomes a candidate revision only when it reflects a reusable analytical improvement rather than a one-off business fact. 

\paragraph{Versioned promotion.}
Candidate revisions from either source do not take effect immediately. They become active only after review, evaluation, and versioning, whether the change adds a new validation signal, refines a planning template, updates a reference, improves a script, or introduces a new workflow. Once promoted, a revision becomes a new skill version available to Host in future tasks, recorded with what changed, why, which evidence or feedback triggered it, how it was evaluated, and how it can be rolled back. This versioned, auditable process lets the method library keep evolving without letting temporary feedback, local preferences, or unsafe rollouts silently change global analytical behavior.

\subsection{What Skill-Hub Guarantees}

By codifying analytical methodology as reusable, hierarchical, and continuously validated assets, Skill-Hub gives \dataaxon three guarantees:
\begin{itemize}
    \item \emph{Reusability}: because method is separated from changing business facts, the same skill applies across domains.
    \item \emph{Methodological consistency}: similar tasks follow the same stable, expert-designed procedures rather than one-off model reasoning.
    \item \emph{Verifiable evolvability}: skills keep improving from feedback and evaluation, but only through candidate revisions that are validated, versioned, and auditable before they take effect.
\end{itemize}
Together, these guarantees let Host execute analytical workflows that are not only grounded by DataBridge but also guided by reusable, continuously improving methodology.

\section{Host Runtime}
\label{sec:host}

\subsection{Role: The Runtime Execution Subsystem}

Host is the runtime execution subsystem of \dataaxon and the only execution subject in its design. DataBridge and Skill-Hub supply governed evidence and reusable methods, while Host materializes these assets into executable analytical runs, turning workflow specifications into concrete runtime actions such as task materialization, subagent dispatch, tool invocation, sandboxed execution, artifact recording, reflection, and failure recovery. By making execution explicit as state, actions, artifacts, and events, it realizes the \emph{Controllable Long-horizon Execution Environment} highlighted in Section~\ref{sec:intro-what}.

\begin{figure}[t]
\centering
\includegraphics[width=0.7\linewidth]{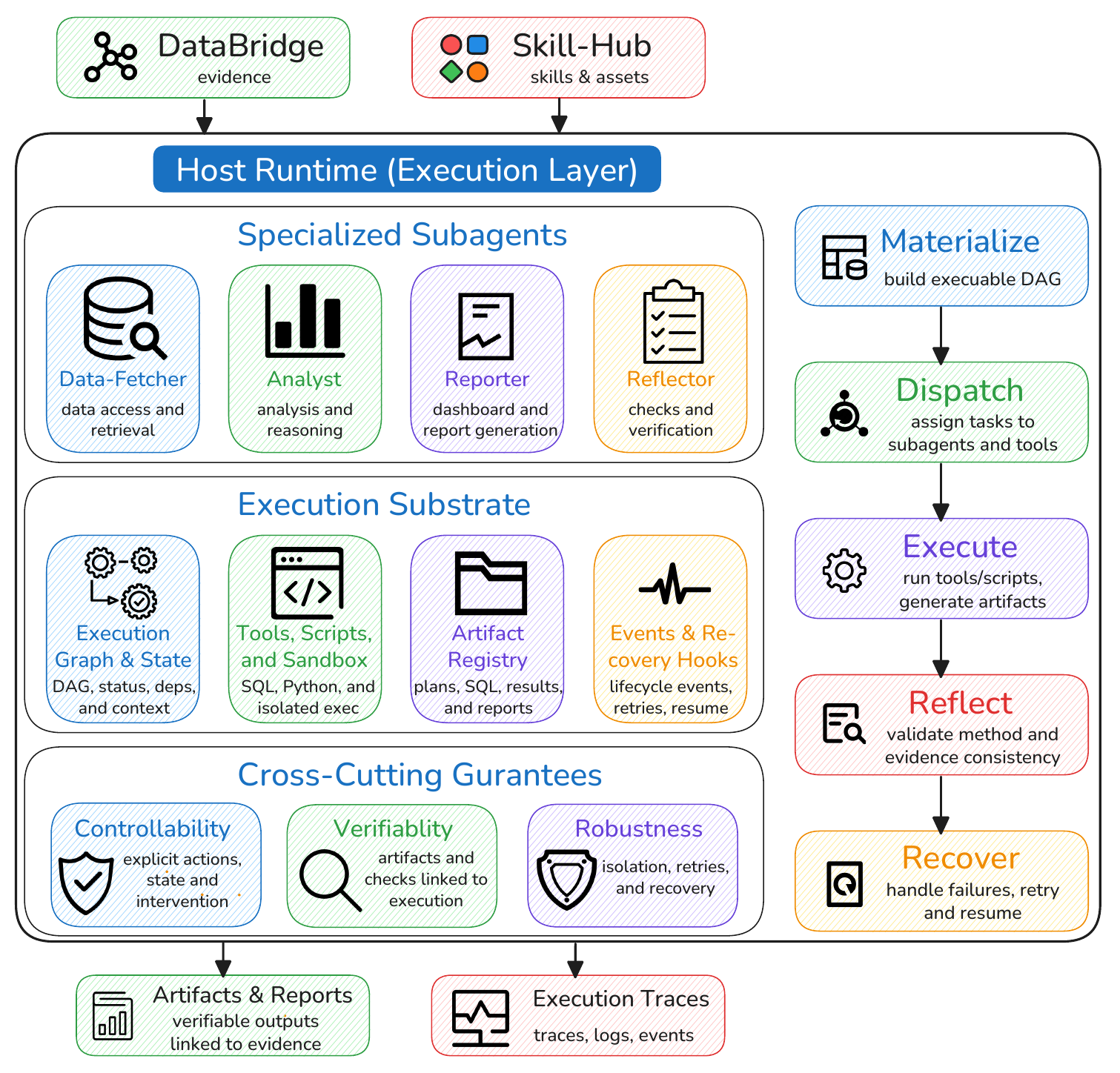}
\caption{The overall structure of Host Runtime. DataBridge supplies governed evidence and Skill-Hub supplies method assets, and Host materializes them through specialized subagents and a managed execution substrate, driven by a runtime lifecycle that makes analytical execution controllable, verifiable, and robust.}
\label{fig:host-runtime}
\end{figure}

As summarized in Figure~\ref{fig:host-runtime}, Host combines a set of specialized subagents and a managed execution substrate (left) with a runtime lifecycle that turns method and evidence assets into executable, inspectable, and recoverable workflows (right). The rest of this section describes the runtime objects Host manages, this lifecycle, and the guarantees it provides.

\subsection{What Host Manages}

Host manages the execution substrate that lets evidence and methods be used safely in real tasks, built from the following runtime objects.

\paragraph{Execution graph and runtime state.}
Host represents each task as a directed acyclic graph (DAG), with executable actions as nodes and their dependencies as edges, a structure expressive enough to capture arbitrarily complex analytical tasks. On top of this graph, Host maintains the task's runtime state, including the status of each action (ready, running, completed, failed, skipped, or awaiting user revision) along with session context, action inputs and outputs, dependency status, retry state, and the current position in the workflow. This explicit representation turns a long-running task from something hidden inside a conversation into a process that is inspectable and easy to manage, recover, and parallelize.

\paragraph{Specialized subagents.}
Host organizes complex analysis through multiple specialized subagents rather than a single monolithic agent. A data-fetching subagent focuses on evidence-aware data access and retrieval. An analysis subagent carries out the core analytical work of a task, such as metric computation, drill-down, attribution, comparison, and constructing intermediate conclusions. A report subagent turns verified findings, charts, and artifacts into an interactive report or dashboard. A reflection subagent uses automatically generated checklists, derived from Skill-Hub assets and task context, to inspect whether the execution process followed the expected method and whether the final answer is supported by artifacts. Dividing work this way gives each subagent a focused context, which keeps long analytical tasks reliable.

\paragraph{Tools, scripts, and sandboxes.}
The actual computation of a task runs through tools and scripts that Skill-Hub supplies or references, covering operations such as SQL queries, Python data processing, visualization, file handling, and report rendering. Rather than running them directly, Host executes these tools and scripts inside isolated sandboxes with their own runtime environments and permission boundaries. This lets a task use flexible computational tools while keeping the host runtime and other sessions safe from unsafe side effects.

\paragraph{Artifact registry.}
Host maintains an artifact registry for the intermediate and final products of execution, including but not limited to plans, SQL statements, query results, transformed datasets, charts, execution logs, generated checklists, reflection results, and final reports. Each artifact links back to the DAG node, tool invocation, and supporting evidence that produced it, so any conclusion can be traced to the concrete operations behind it. Because artifacts persist in the registry, they can also be referenced and reused across chat sessions instead of being regenerated, which improves efficiency, especially for the results of large-scale data retrieval.

\paragraph{Runtime events and recovery hooks.}
Host emits runtime events at key execution stages such as tool invocation, artifact generation, reflection, and failure handling, and exposes lifecycle hooks on them. Combined with the per-node state recorded in the DAG, these hooks enable recovery and user intervention. At a hook, Host can preserve completed upstream artifacts and retry, roll back, or resume the affected nodes from a checkpoint instead of restarting the task, for example when a query times out or a user revises the plan mid-run.

\subsection{Runtime Execution Lifecycle}

Host runs each analytical task through an execution lifecycle of five stages that turn method and evidence assets into a controlled, inspectable execution process:

\textbf{Materialize.} Host first interprets the selected Skill-Hub workflow specification in light of the user request and the evidence retrieved from DataBridge, then expands its procedure steps into concrete actions, wires up the dependencies among them, and determines the inputs and intermediate artifacts each action needs. The result is an executable DAG for the task.

\textbf{Dispatch.} Host then schedules the ready actions and routes each to the subagent or tool that fits its type, so data-fetching, analytical, report- generation actions go to their corresponding subagents while independent branches run in parallel once their dependencies are satisfied.

\textbf{Execute.} The dispatched actions run as tool calls and script execution inside sandboxes, and every product they generate, such as SQL statements, query results, transformed datasets, charts, logs, and draft reports, is written into the artifact registry and linked back to its DAG node.

\textbf{Reflect.} As actions complete, Host invokes the reflection subagent, which runs against a task-specific checklist automatically generated from the relevant Skill-Hub specification, references, and expected artifacts, checking whether the retrieved evidence, stated assumptions, intermediate artifacts, selected dimensions, and final conclusions are consistent with the intended method.

\textbf{Recover.} When an action fails, is interrupted, or is revised by the user, Host uses runtime hooks and persisted state to retry, branch, roll back, or resume the affected actions from a checkpoint, without invalidating completed upstream artifacts.

\subsection{What Host Guarantees}

By turning method and evidence into execution that is explicit, traceable, and recoverable, Host gives \dataaxon three guarantees:
\begin{itemize}
    \item \emph{Controllability}: execution unfolds as explicit actions, dependencies, state, and intervention points rather than as an opaque model conversation, so a task can be inspected and steered at any stage.
    \item \emph{Verifiability}: intermediate artifacts, tool calls, generated checklists, and reflection results all link back to the execution graph, so every conclusion can be traced to the evidence and operations that produced it.
    \item \emph{Robustness}: sandbox isolation, persisted state, and checkpoint-based recovery keep a local tool, script, or stream failure from collapsing the whole workflow.
\end{itemize}
Together, these guarantees let \dataaxon execute open-form data-analysis tasks as durable, inspectable, and recoverable workflows, grounded by DataBridge and guided by Skill-Hub.
\section{Evaluation}
\label{sec:eval}

Having presented the design of \dataaxon, we now examine whether it delivers measurable benefits in practice. Our evaluation is centered on two questions:

\begin{itemize}
    \item \emph{Q1. To what extent does \dataaxon improve over a general-purpose agent in real enterprise analysis?}
    \item \emph{Q2. How does each of its components contribute to these gains?}
\end{itemize}

To investigate these questions under realistic conditions rather than in a sandbox, we deployed \dataaxon within an internal business line, where it serves business-intelligence (BI) users in their day-to-day analytical work and supplies the workloads used throughout this evaluation. There, analysts issue genuine business requests over live enterprise data hosted in Hologres, whose heterogeneity, ambiguity, and continual evolution reflect the difficulties discussed in Section~\ref{sec:intro-diff}.

\subsection{Enterprise Gains over a General Agent}
\label{sec:eval-deploy}

To answer the first question, we evaluate \dataaxon on tasks drawn from the BI users' daily usage, which fall into two categories. The first consists of 29 objective queries, such as NL2SQL and metric computation, each with an exact ground-truth answer. The second consists of 37 distinct open-ended analytical queries, spanning business metric monitoring and anomaly diagnosis, trend and growth analysis, periodic business reporting, and ad-hoc deep-dive attribution; each is a long-horizon task that requires from roughly 15 minutes to one hour of autonomous execution and culminates in a report.

\paragraph{Objective queries.} We judge each objective query by exact agreement with a reference answer that has been reviewed by more than two domain experts. On these queries, \dataaxon returns the correct result on about 96.5\% of the cases. The main reason is that DataBridge grounds each query in the right metrics, dimensions, and tables, so the agent binds to the correct entities instead of guessing, while Skill-Hub contributes two assets: a robust data-retrieval pipeline that chains query rewriting, context gathering, disambiguation, SQL generation, and error recovery; and standardized metric-computation scripts. Together they let the agent follow a verified procedure rather than ad-hoc logic.

\paragraph{Open-ended queries.} For each open-ended query, we let business users give a satisfaction score on a 100-point scale after they have used the result. Under this measure, the deployed \dataaxon attains an average satisfaction of 78.3. When we replace it with a leading general-purpose agent from the strongest tier of current systems, one that lacks DataBridge and Skill-Hub, satisfaction drops to 34.1. To keep the comparison fair, both systems are driven by the same underlying LLM with fixed request parameters, so the reported gain is a conservative lower bound attributable purely to the harness.
The gap traces directly to \dataaxon's three components. DataBridge grounds each query in the right metrics and dimensions, Skill-Hub supplies the expert analytical procedure, and the Host runtime keeps the long-horizon, multi-step execution reliable and every artifact traceable, so \dataaxon tends to deliver a complete analytical loop: it presents the overall trend, abnormal fluctuations, and drill-down attribution layer by layer, and keeps every visualization consistent with the stated conclusion. The general-purpose agent, without grounded semantics or a codified procedure, guesses at the semantics and skips the deeper analytical steps, so its answers drift and stay shallow. 

As an additional benefit, \dataaxon brings two further gains. First, it annotates data lineage and query provenance throughout, so every result is fully auditable, which is essential in high-stakes data analysis where every conclusion must be traced back to its source before it can drive a business decision. Second, rather than leaving the LLM to explore the raw schema on its own, which is extremely costly when an industrial warehouse holds vast numbers of tables and columns, DataBridge supplies exactly the metrics, dimensions, and table context each step needs. In our measurements, this reduces context token consumption by roughly 42\% on average.

\subsection{Component Ablation}
\label{sec:eval-ablation}

Having seen the end-to-end gains, we now answer the second question by isolating how each added component contributes. Holding the same underlying LLM fixed, we start from the general-purpose agent and enable Skill-Hub and DataBridge, individually and then together as the full \dataaxon, on the same open-ended requests. To evaluate these configurations at scale under a single consistent standard, we ask the BI users to express their expectations for each request as explicit checkpoints along four dimensions, and an LLM-as-a-judge then scores every configuration against these checkpoints. The dimensions capture what business users care about: \emph{analysis breadth}, whether the analysis covers the relevant metrics, dimensions, and segments rather than a single angle; \emph{analysis depth}, whether it moves beyond surface numbers to anomaly attribution, root-cause decomposition, and actionable conclusions; \emph{report quality}, whether the derived report is clear, well organized, and has every conclusion supported by data and visualizations; and \emph{artifact completeness}, whether the expected artifacts, such as the analysis plan, query results, intermediate computation results, Python code, and charts, are all persisted to disk and thus traceable. Table~\ref{tab:ablation} reports the results.

\begin{table}[t]
\centering
\caption{Component ablation on production BI tasks. All scores are percentages; higher is better.}
\label{tab:ablation}
\begin{tabular}{lcccc}
\toprule
\textbf{Configuration} & \makecell{\textbf{Analysis}\\\textbf{Breadth}} & \makecell{\textbf{Analysis}\\\textbf{Depth}} & \makecell{\textbf{Report}\\\textbf{Quality}} & \makecell{\textbf{Artifact}\\\textbf{Completeness}} \\
\midrule
General-purpose agent & 27.35 & 25.21 & 35.64 & 36.94 \\
\quad + Skill-Hub & 66.32 & 48.46 & 47.03 & 85.90 \\
\quad + DataBridge & 36.59 & 29.39 & 37.84 & 32.43 \\
\dataaxon (both) & \textbf{79.63} & \textbf{62.60} & \textbf{58.09} & \textbf{86.04} \\
\bottomrule
\end{tabular}
\end{table}

The general-purpose agent alone stays low across all four dimensions, consistent with the low satisfaction it received in the deployment comparison of Section~\ref{sec:eval-deploy}. Adding Skill-Hub sharply lifts analysis breadth and artifact completeness: it supplies carefully curated analytical skills that walk the agent through the complete analysis flow layer by layer, while its runtime skills enforce persistence conventions that write every intermediate artifact to disk and keep it traceable. Yet analysis depth and report quality stay limited, because without grounded metric and dimension semantics and the relationships among them, the agent is unsure of the right direction and its analysis can drift, for example failing to drill down into the secondary or even tertiary dimensions, which in turn leaves the report thin. Adding DataBridge alone grounds the semantics but yields only modest gains, because without a codified methodology the agent skips the complete analytical steps. Only with both does \dataaxon reach the top level across all four dimensions, and the combined gain clearly exceeds the sum of the two components applied alone, showing that grounding evidence and methodology are not merely complementary but mutually reinforcing, delivering their full value only when present together. 

\subsection{Reproducibility on Open-Source Benchmarks}
\label{sec:eval-public}

The evaluations in Sections~\ref{sec:eval-deploy} and~\ref{sec:eval-ablation} run on proprietary enterprise workloads that cannot be publicly disclosed. To let others verify \dataaxon's underlying capability in a reproducible and comparable setting, we also report results on two public data-agent benchmarks, KramaBench~\citep{kramabench} and DAComp~\citep{dacomp}, which echo different facets of enterprise analysis. KramaBench builds end-to-end data analysis pipelines over heterogeneous, multi-source data lakes, which requires sifting the useful evidence out of a large volume of raw data, and thus mirrors the data part of our real tasks. DAComp instead mimics the data-intelligence lifecycle over larger databases and calls for more open-ended reasoning, closer to the intent and scale of real analytical queries. Both go beyond a single lookup and require multi-step pipelines from raw data to insight.

\begin{table}[t]
\centering
\caption{Results on public data-agent benchmarks. Higher is better; each column follows the benchmark's official metric. The Human and current SOTA figures are taken from the respective benchmark papers~\citep{kramabench,dacomp}.}
\label{tab:public-bench}
\begin{tabular}{ccc}
\toprule
\textbf{Method} & \textbf{KramaBench} & \textbf{DAComp} \\
\midrule
Human & 76.75 & -- \\
Current SOTA & 55.83 & 50.84 \\
\dataaxon & 68.32 & 62.38 \\
\bottomrule
\end{tabular}
\end{table}

As shown in Table~\ref{tab:public-bench}, \dataaxon leads the current state of the art on both benchmarks, and on KramaBench it closes much of the remaining gap to human experts. It reaches this with the same design that drives its enterprise gains: through a standardized data-retrieval and discovery pipeline it narrows a sprawling data lake to the most useful information, and it keeps the multi-step analysis on a disciplined path from raw data to insight. That this lead holds on public data confirms that the underlying capability generalizes beyond our private workloads.

\paragraph{Summary.} Overall, these results answer the two questions posed at the outset. In real enterprise analysis, \dataaxon resolves objective queries with high accuracy and substantially outperforms a strong general-purpose agent on open-ended tasks (Q1); this advantage stems from the interplay between grounded enterprise semantics and codified analytical skills, neither of which suffices in isolation (Q2); and its lead on public benchmarks shows that the capability is reproducible and generalizes beyond our own deployment.

\section{Future Work}

The current version of \dataaxon establishes the core architecture for agentic data analysis, with DataBridge grounding execution in semantic evidence, Skill-Hub standardizing reusable analytical methods, and Host turning both into controllable runtime workflows. Building on this foundation, future work follows the same architectural line and moves from a reliable analysis framework toward a continuously improving enterprise data intelligence system, along three directions.

\paragraph{Strengthening the semantic--method--runtime flywheel.}
The first direction is to keep advancing each of the three components. DataBridge will pursue richer semantic mining, automatic graph refinement, and semantic evolution from execution traces and user feedback. Skill-Hub will focus on cross-domain skill transferability, so cold-starting a new domain no longer means building analytical methods from scratch. Host will improve execution efficiency, deployment flexibility, and interoperability with emerging agent runtimes. Together, these advances tighten the semantic--method--runtime flywheel, making the framework more reusable across domains and more adaptive to real enterprise usage.

\paragraph{Improving interactive and self-verifying execution.}
The second direction is to make long-running analysis inspectable and correctable while it runs, not only after it finishes. \dataaxon will support stronger human-in-the-loop intervention, letting users inspect intermediate artifacts, revise execution plans, correct analytical procedures, and steer subsequent actions without restarting the workflow. It will also strengthen self-judging mechanisms that estimate result confidence by combining semantic consistency, execution evidence, reflection outcomes, and historical feedback, turning verification from a final manual review into a runtime capability.

\paragraph{Scaling toward enterprise data intelligence.}
The third direction is to extend \dataaxon from an analytical framework toward a data brain for enterprise decision-making and data management, making insight discovery, action recommendation, workflow coordination, and data-asset optimization more systematic and governed at enterprise scale. Supporting this shift will require stronger tenant isolation, fine-grained permission control, secure execution environments, and enterprise-grade governance for large-scale multi-tenant deployments.

\section{Conclusion}

In this report, we presented \dataaxon, an end-to-end framework for agentic data analysis organized around three AI-native architectural questions: \emph{what facts to use}, \emph{how to analyze}, and \emph{how to run the resulting workflow}. DataBridge answers the first with governed semantic evidence that stays grounded and fresh, Skill-Hub answers the second with reusable and consistent analytical methods, and Host answers the third with a runtime that makes execution controllable, verifiable, and robust. Together, these components address the core difficulty of enterprise data analysis, where a system must ground business concepts, follow consistent analytical procedures, and execute long-running workflows in a controllable environment. This modular design offers a practical foundation for continuously improving data-analysis agents and scaling them toward governed, enterprise-level data intelligence.

\nocite{*}

\bibliographystyle{plainnat}
\bibliography{ref}
\end{document}